\documentclass[10pt,twocolumn,letterpaper]{article}

\usepackage[pagenumbers]{cvpr} 
\usepackage{multirow}
\usepackage{booktabs}
\usepackage{colortbl}
\usepackage{xcolor}
\usepackage{makecell}
\usepackage{graphicx}
\usepackage{overpic}
\usepackage{subcaption}
\usepackage[accsupp]{axessibility}
\usepackage[ruled,vlined]{algorithm2e}  
\usepackage{amsmath,amssymb}            
\definecolor{cvprblue}{rgb}{0.21,0.49,0.74}
\usepackage[pagebackref,breaklinks,colorlinks,allcolors=cvprblue]{hyperref}

\def\paperID{} 
\def\confName{CVPR}
\def\confYear{2026}

\title{FlowPalm: Optical Flow Driven Non-Rigid Deformation for \\Geometrically Diverse Palmprint Generation}

\author{
    Yuchen Zou$^{1}$ \quad
    Huikai Shao$^{1,2\ast}$ \quad
    Lihuang Fang$^{3}$ \quad
    Zhipeng Xiong$^{1}$ \quad
    Dexing Zhong$^{1}$\thanks{Corresponding authors.} \\[0.3em] 
    \small $^{1}$Xi'an Jiaotong University \\
    \small $^{2}$Sichuan Digital Economy Industry Development Research Institute \\
    \small $^{3}$Southern University of Science and Technology
     \\[0.3em] 
    {\ttfamily\small yuchenzou@stu.xjtu.edu.cn} 
}

\begin{document}
\maketitle


\begin{abstract}
Recently, synthetic palmprints have been increasingly used as substitutes for real data to train recognition models. To be effective, such synthetic data must reflect the diversity of real palmprints, including both style variation and geometric variation. However, existing palmprint generation methods mainly focus on style translation, while geometric variation is either ignored or approximated by simple handcrafted augmentations. In this work, we propose FlowPalm, an optical-flow-driven palmprint generation framework capable of simulating the complex non-rigid deformations observed in real palms. Specifically, FlowPalm estimates optical flows between real palmprint pairs to capture the statistical patterns of geometric deformations. Building on these priors, we design a progressive sampling process that gradually introduces the geometric deformations during diffusion while maintaining identity consistency. Extensive experiments on six benchmark datasets demonstrate that FlowPalm significantly outperforms state-of-the-art palmprint generation approaches in downstream recognition tasks. Project page: \url{https://yuchenzou.github.io/FlowPalm/}
\end{abstract}

\section{Introduction}

\begin{figure}[t]
    \centering
    \begin{subfigure}[b]{0.98\columnwidth}
        \centering
        \includegraphics[width=\columnwidth]{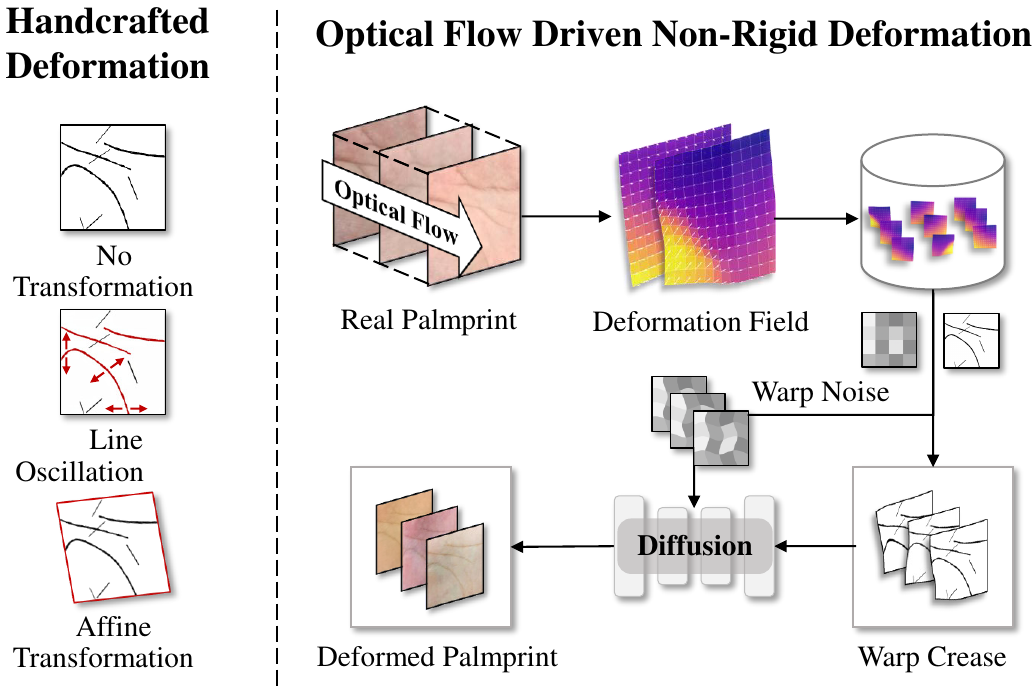} 
        \put(-187,88){\scriptsize \cite{jin2025diff}}
        \put(-198,47){\scriptsize \cite{shen2023rpg, zou2025pfig}}
        \put(-187,5){\scriptsize \cite{jin2024pce}}
        \caption{Comparison of deformation schemes adopted by different palmprint generation methods.}
        \label{fig:sub1}
    \end{subfigure}
    \hfill
    \begin{subfigure}[b]{0.82\columnwidth}
        \centering
        \includegraphics[width=\columnwidth]{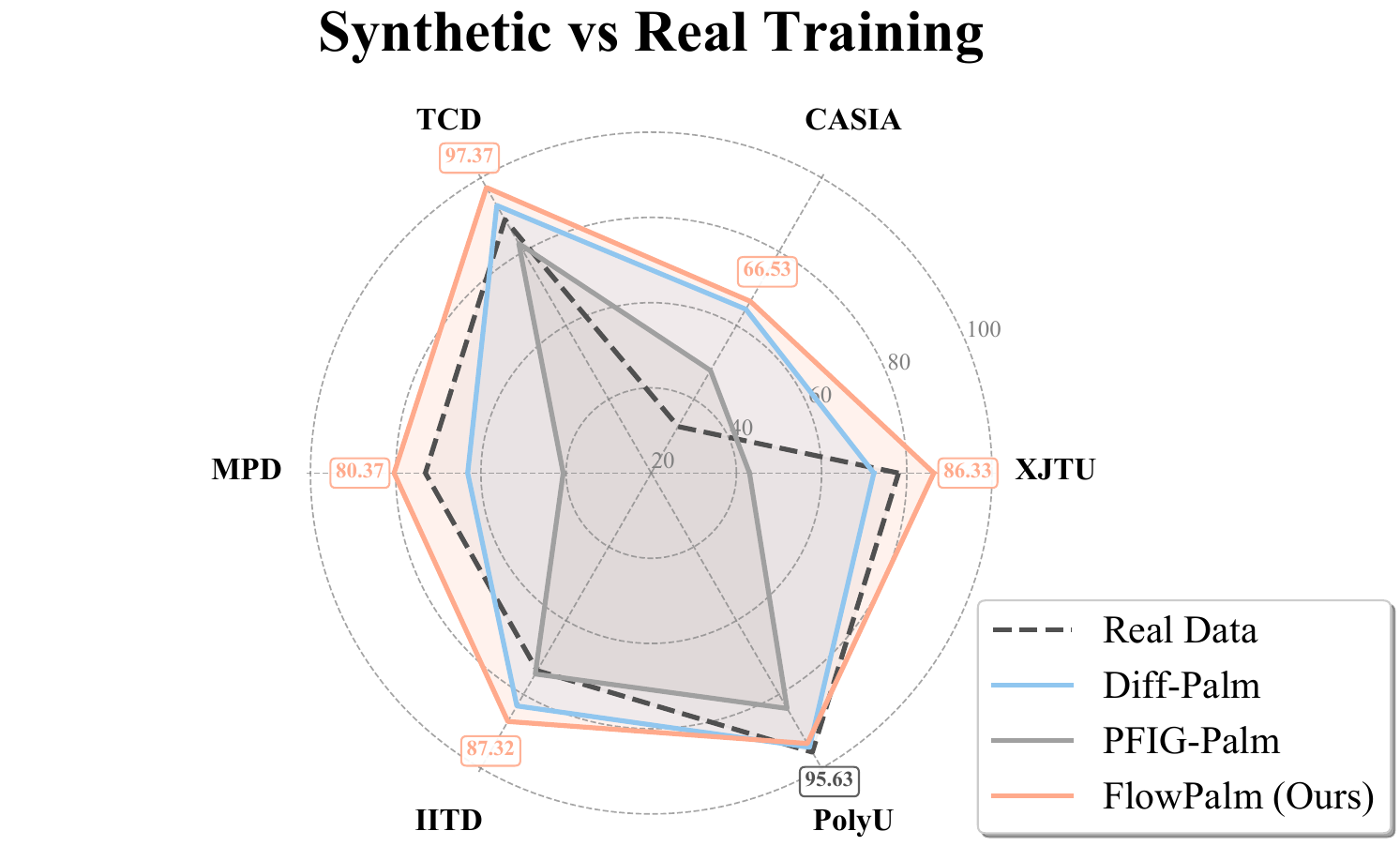} 
        \put(-17,20){\scriptsize \cite{jin2025diff}}
        \put(-15,13){\scriptsize \cite{zou2025pfig}}
        \caption{Quantitative comparison of palmprint generation methods}
        \label{fig:sub2}
    \end{subfigure}
    \caption{Overview of the proposed method and its performance comparison with existing methods. (a) We introduce optical flow driven non-rigid deformation modeling to overcome the limitations of handcrafted simple transformations, enabling the synthesis of geometrically diverse and identity-consistent palmprints. (b) Experimental results on six public databases demonstrate that FlowPalm achieves superior recognition performance.}
    \label{fig1}
\end{figure}
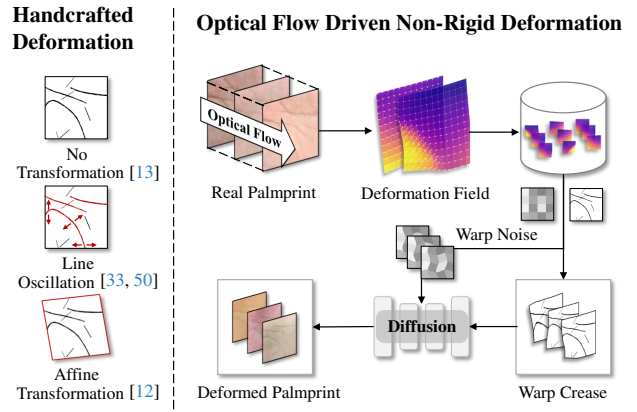
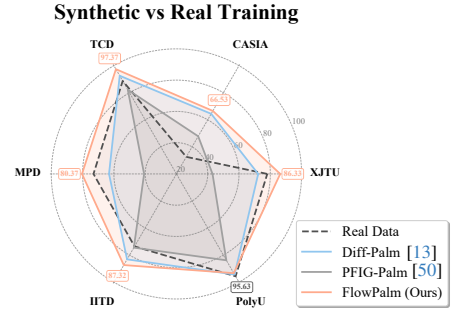

Palmprint recognition has emerged as a prominent biometric modality due to its rich textural patterns and strong privacy protection~\cite{yang2024physics}. With the rapid advancement of deep learning technologies, neural network-based palmprint recognition models have demonstrated remarkable robustness and discriminative power~\cite{fan2023amgnet, zhao2022structure, liu2025sf2net}. However, the training of these high-performance models heavily relies on large-scale, diverse, and high-quality datasets~\cite{pan2025hierarchical, shao2024generating}. In real-world scenarios, the acquisition of such datasets is often hindered by privacy protection policies and strict acquisition protocols~\cite{liu2022data}.

To alleviate this problem, recent research has explored the use of generative models such as Generative Adversarial Networks (GANs)~\cite{goodfellow2020gan} and diffusion models~\cite{ho2020denoising} to synthesize palmprint images with novel identities. These synthetic palmprints typically generated through line-based condition~\cite{zhao2022bezierpalm}, have proven effective in enhancing recognition performance while reducing dependence on real data.

Despite recent progress in palmprint synthesis, existing methods are largely limited to appearance modeling. They primarily enhance styles or transfer textures, while paying insufficient attention to the geometric variability inherent in real palmprints. In unconstrained environments, palmprint images naturally undergo complex deformations induced by hand bending, finger articulation, camera viewpoint variation, and imaging parameter differences~\cite{Zhong2025RegPalm}.
These deformations represent a crucial dimension of the palmprint distribution that determines recognition robustness. Unfortunately, existing methods either completely ignore geometric deformation (e.g., Diff-Palm~\cite{jin2025diff}) or employ simplified handcrafted perturbations, such as applying line oscillation (RPG-Palm~\cite{shen2023rpg}, PFIG-Palm~\cite{zou2025pfig}) or affine transformation (PCE-Palm~\cite{jin2024pce}). As a result, the synthetic data lack realistic and diverse geometric transformations, leading to a performance bottleneck in palmprint recognition.

Generating palmprints with realistic deformations is therefore far from trivial. It poses two main challenges: (1) accurately modeling and simulating the complex, spatially varying non-rigid geometric deformations of real palms, and (2) preserving identity consistency while introducing such deformations.

To address these challenges, we propose \textbf{FlowPalm}, an optical flow driven palmprint generation framework capable of synthesizing geometrically diverse yet identity-consistent palmprint images. As illustrated in~\cref{fig1}, we first estimate dense optical flow fields~\cite{teed2020raft} between pairs of real palmprint images to capture the non-rigid deformation patterns of real palms. To ensure the reliability of these deformation fields, we further design a palm deformation evaluation pipeline to filter out invalid or abnormal optical flow results. The high-quality deformation fields are then organized into a deformation library, serving as a geometric prior for the subsequent generation process.

To incorporate the deformation prior while maintaining identity consistency, we design a three-stage progressive generation strategy: deformed principal line generation, deformed texture generation, and unconditional texture refinement. 
In the first stage, deformation fields sampled from the library warp the creases and guide the diffusion model, followed by a clean denoising step to obtain clear principal lines. In the second stage, correspondingly warped noise is injected into these clean line maps to synthesize textures that are both geometrically deformed and identity-consistent. In the third stage, to prevent over-constrained texture details caused by manual crease conditions, we remove the crease input and adopt an unconditional denoising process to further refine texture realism.

Based on this deform-guided generation framework, FlowPalm can further generate dynamic palmprint transformation videos that intuitively visualize continuous non-rigid deformations, as shown in the supplementary material.

The main contributions of this work are as follows:
\begin{itemize}
    \item We propose a novel optical flow driven palmprint generation framework called FlowPalm, which synthesizes palmprints with complex non-rigid deformations while preserving identity consistency.

    \item We leverage optical flow estimation to statistically analyze and filter real deformations as reliable priors. Based on this, we design a deformation-driven generation strategy that progressively generates deformed principal line and textures, where the removal of manual crease condition in final stage significantly enhances realism.

    \item Extensive evaluations across six benchmark datasets demonstrate that models trained exclusively on FlowPalm-generated deformed data surpass real data, while further achieving state-of-the-art (SOTA) verification performance across various training paradigms.
\end{itemize}

\section{Related Work}

\subsection{Data Generation for Recognition Tasks}

To alleviate data scarcity in biometric recognition, researchers have explored various approaches for synthetic data generation~\cite{joshi2024synthetic}. Generative Adversarial Networks (GANs) and diffusion models have been widely employed to augment face datasets~\cite{papantoniou2024arc2face,xu2024id,boutros2023idiff,wang2025facea}. These methods often sample identity embeddings extracted by pretrained recognition models~\cite{deng2019arcface,kim2022adaface} to control identity information during the generation process.

In the domain of palmprint synthesis, early works attempted to generate palmprint images using unconditional GANs~\cite{shao2019palmgan,shao2021jpfa,zhu2023contactless}, but lacked the ability to create and manipulate new identities. A milestone work by Zhao et al.~\cite{zhao2022bezierpalm} parameterized palm crease patterns using Bézier curves, enabling the creation of numerous synthetic identities. Motivated by this, several studies have adopted curve-based representations to generate identity-controllable palmprints, including RPG-Palm~\cite{shen2023rpg}, PCE-Palm~\cite{jin2024pce}, and PFIG-Palm~\cite{zou2025pfig} based on GANs, as well as Diff-Palm~\cite{jin2025diff} based on diffusion models.

However, existing approaches mainly focus on enriching style diversity, while the simulation of geometric deformation in palmprints remains relatively simple. Some methods completely ignore geometric deformation~\cite{jin2025diff}, while others apply only simplified perturbation strategies, such as line oscillation~\cite{shen2023rpg,zou2025pfig} or affine transformations~\cite{jin2024pce, jin2025unified}. These approaches fail to reproduce the complex non-rigid deformations that naturally occur in real palms, resulting in synthetic palmprints with limited geometric diversity.

\subsection{Condition and Noise in Diffusion Models}
Denoising Diffusion Probabilistic Models (DDPMs)~\cite{ho2020denoising} learn a reverse diffusion process that progressively denoises Gaussian noise to generate realistic data. Researchers have introduced various conditioning mechanisms (such as class labels~\cite{dhariwal2021diffusion}, text prompts~\cite{nichol2022glide,rombach2022ldm}, and structural priors~\cite{zhang2023controlnet,koley2024sketch,koley2024text}) to achieve controllable generation. These conditional signals are typically encoded and injected into the denoising network, guiding the reverse process toward samples that satisfy the desired semantics.

Recent studies have revealed that diffusion models exhibit remarkable spatial consistency between the noise and content spaces~\cite{chang2024warp}. Specifically, spatial transformations applied to the input noise can lead to corresponding structural variations in the generated images, reflecting an inherent equivariance between noise and image domains~\cite{zhou2025alias, liu2025equivdm}. This property provides a new perspective for controllable generation: the generative process can be influenced not only through explicit conditioning signals but also by directly manipulating the input noise to control the geometric shape~\cite{muller2024multidiff, yan2025consistent} or spatial layout~\cite{burgert2025go} of the generated content.

Leveraging this noise–content equivariance, recent palmprint generation studies~\cite{jin2025diff} have employed identical noise samples during the late diffusion stages to enforce texture consistency across generated palmprints. Inspired by both conditional control and noise manipulation, we propose to jointly deform the crease condition and noise input during diffusion. In the early stage, a geometric distortion is applied to the crease condition (representing the principal line structure) to control the global deformation, followed by clean denoising to remove stochastic noise. In the later stage, noise with a distortion consistent with the principal lines is injected to ensure texture-level deformation consistency. Experiments demonstrate that this joint deformation strategy effectively maintains identity consistency between the principal lines and fine-grained textures under non-rigid deformation.

\section{Method}
\begin{figure}
    \centering
    \includegraphics[width=0.95\linewidth]{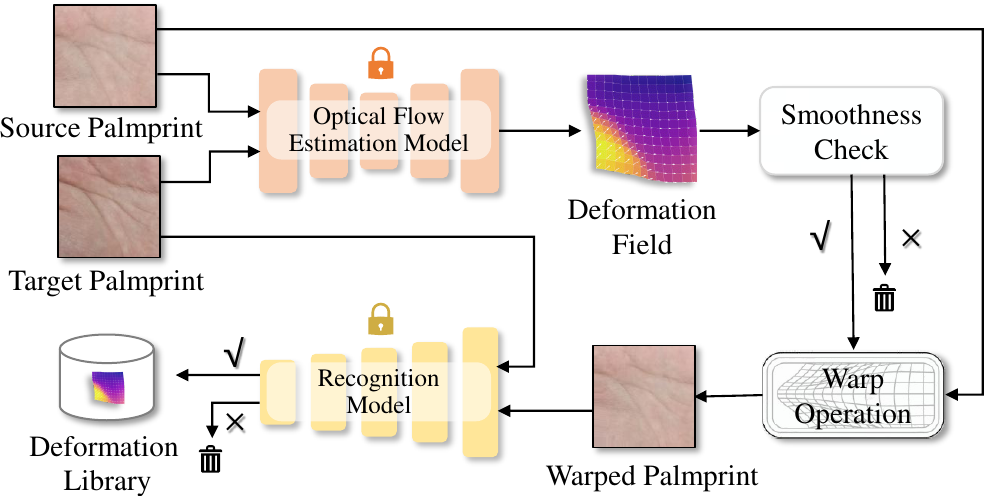}
    \caption{
    Overview of the deformation prior construction. Deformation field is estimated between real palmprint pairs and filtered through smoothness and identity checks to build a reliable deformation library for generation.
    }
    \label{fig:flow_frame}
\end{figure}

\begin{figure}
    \centering
    \includegraphics[width=0.95\linewidth]{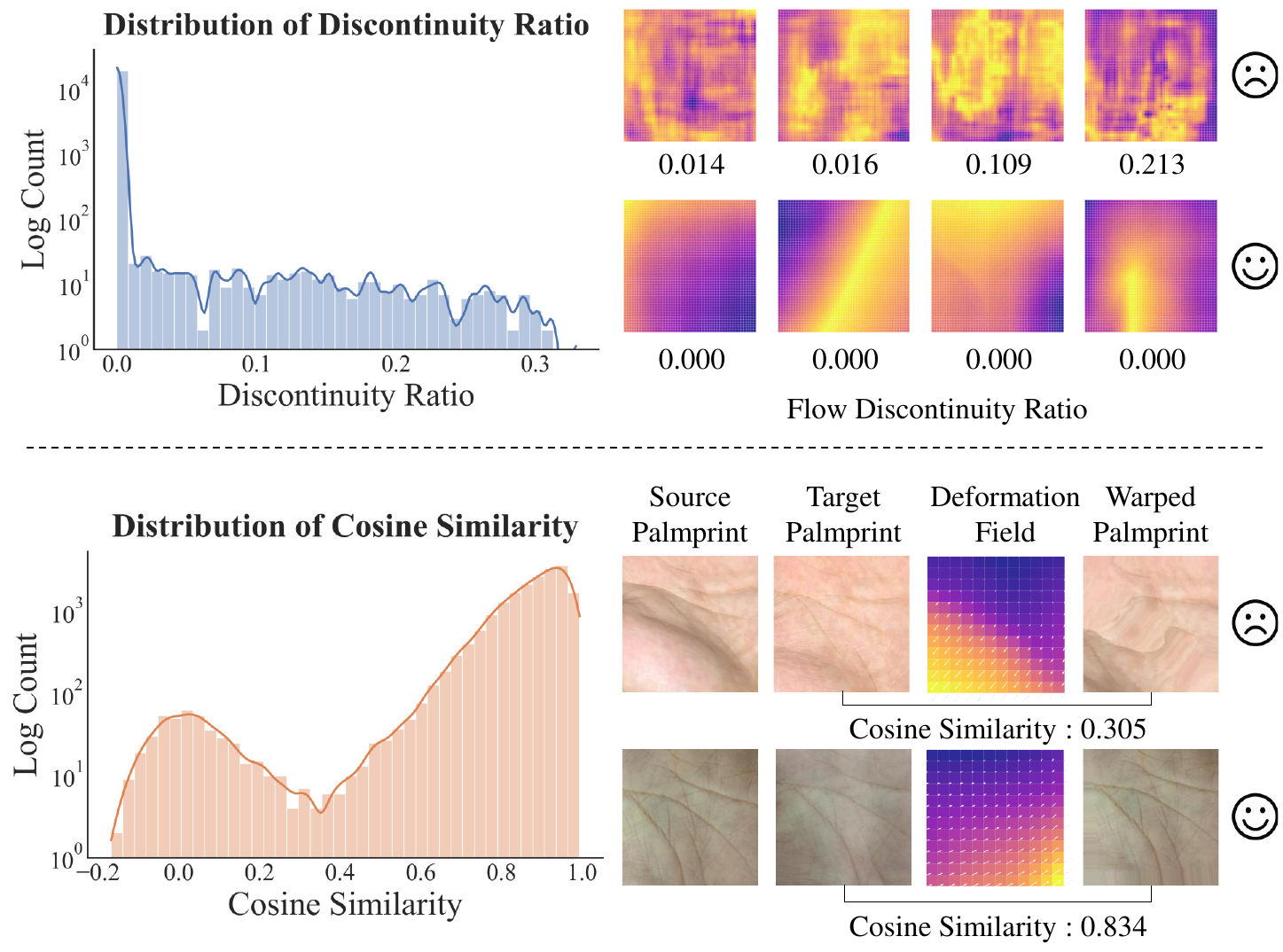}
    \caption{Statistical distribution and visualization of deformation quality assessment.}
    \label{fig:flow_visual}
\end{figure}

\begin{figure*}
    \centering
    \includegraphics[width=0.88\linewidth]{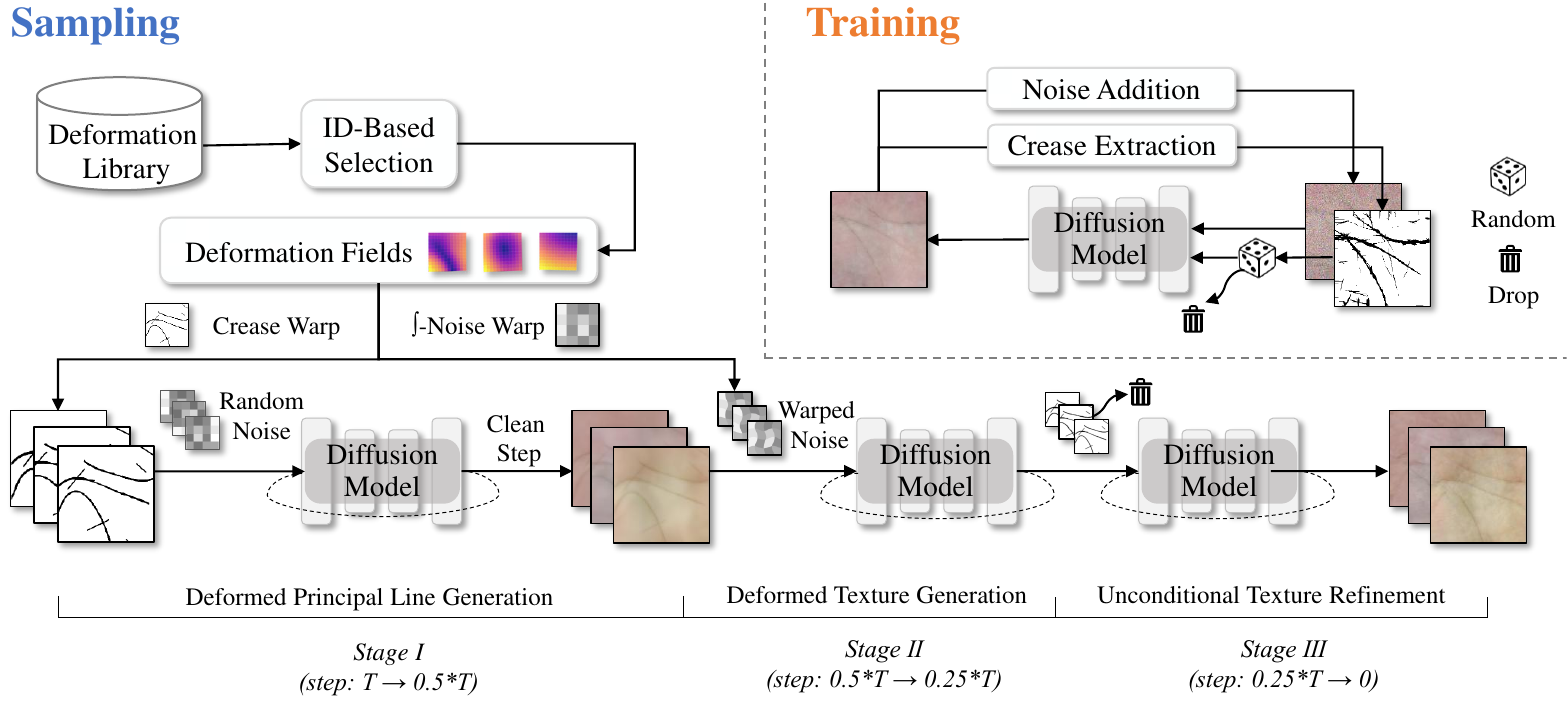}
    \caption{Overview of the proposed deformation-driven three-stage generation strategy. 
A deformation field sampled from the deformation library guides the synthesis process: 
Stage~I generates deformed and appearance-diverse principal lines under warped crease conditions and random noise; Stage~II produces identity-consistent deformed textures under warped crease conditions and warped crease noise; and Stage~III refines texture details through unconditional denoising.}
    \label{fig:total_frame}
\end{figure*}

We propose \textbf{FlowPalm}, an optical flow–driven palmprint generation framework designed to synthesize geometrically diverse yet identity-consistent palmprint images. 
The overall pipeline consists of two main components: 
(1) a \emph{deformation prior construction} module that estimates and filters non-rigid optical flow fields from real palmprint pairs (illustrated in~\cref{fig:flow_frame}), and 
(2) a \emph{deformation-driven palmprint generation} module that progressively synthesizes deformed principal lines and textures under the guidance of the deformation prior (illustrated in~\cref{fig:total_frame}). In the following subsections, we describe these two components in detail.

\subsection{Deformation Prior Construction}

As illustrated in~\cref{fig:flow_frame}, the goal of deformation prior construction is to statistically capture realistic non-rigid geometric variations observed in real palmprints. Given a pair of palmprint images from the same identity, denoted as $\mathbf{I}_s$ (source) and $\mathbf{I}_t$ (target), we employ a pretrained optical flow estimation model (RAFT~\cite{teed2020raft}) to estimate the dense correspondence between them. The model outputs a deformation field $\mathbf{F} = (u, v)$, where each vector $(u_{x,y}, v_{x,y})$ represents the horizontal and vertical displacements of pixel $(x, y)$ from $\mathbf{I}_s$ to $\mathbf{I}_t$:
\begin{equation}
    \mathbf{I}_t(x, y) \approx \mathbf{I}_s(x + u_{x,y},\, y + v_{x,y}), 
    \quad \mathbf{F} \in \mathbb{R}^{H \times W \times 2}.
\end{equation}

These deformation fields inherently capture both local elastic bending and global shape transformations induced by palm and finger pose changes. However, not all estimated flow fields are reliable; some may exhibit discontinuities, noisy vectors, or invalid correspondences. To construct a robust deformation prior, we perform a two-level quality assessment procedure.

\paragraph{Smoothness Validation.} 
A physically plausible palm deformation should vary smoothly within local neighborhoods and avoiding abrupt discontinuities. 
We quantify the discontinuity ratio of the deformation field $\mathbf{F}$ as:
\begin{equation}
    \mathcal{D}(\mathbf{F}) = 
    \frac{1}{HW} 
    \sum_{x,y} 
    \mathbb{I}\!\left(
    \|\nabla \mathbf{F}(x, y)\|_2 > \delta
    \right),
\end{equation}
where $\nabla \mathbf{F}$ denotes the spatial gradient of the flow field, and $\delta$ is empirically set to $5$. A deformation field is considered smooth if $\mathcal{D}(\mathbf{F}) < \tau_d$, where $\tau_d$ is an empirically determined threshold. Lower discontinuity ratios indicate smoother and more physically reasonable deformation patterns, as illustrated in the upper part of~\cref{fig:flow_visual}.

\paragraph{Identity Consistency Check.}
Even when geometrically plausible, a failed deformation often manifests as distortions of identity-related texture patterns. Therefore, we further evaluate whether the warped palmprint $\hat{\mathbf{I}}_t$ produced by $\mathbf{F}$ can still be recognized as the same subject by a pretrained recognition model $\mathcal{R}(\cdot)$. 
The warping operation is defined by bilinear sampling as:
\begin{equation}
\mathcal{W}(\mathbf{C}, \mathbf{F})(x)
= \sum_{i \in \mathbb{Z}^2} \mathbf{C}(i)\, k(x+\mathbf{F}(x)-i),
\label{eq:warp}
\end{equation}
where $k(\cdot)$ denotes the bilinear interpolation kernel. 
Accordingly, the warped palmprint is expressed as:
\begin{equation}
\hat{\mathbf{I}}_t = \mathcal{W}(\mathbf{I}_s, \mathbf{F}),
\end{equation}
and the cosine similarity between $\hat{\mathbf{I}}_t$ and $\mathbf{I}_t$ features is computed as:

\begin{equation}
    \mathcal{C}(\mathbf{F}) = 
    \frac{\left\langle \mathcal{R}(\hat{\mathbf{I}}_t),\, \mathcal{R}(\mathbf{I}_t) \right\rangle}
    {\| \mathcal{R}(\hat{\mathbf{I}}_t) \|_2 \cdot \| \mathcal{R}(\mathbf{I}_t) \|_2}.
\end{equation}

Only deformation fields satisfying $\mathcal{C}(\mathbf{F}) > \tau_c$ are retained. 
Examples of poor and high-consistency deformations are presented in the lower part of~\cref{fig:flow_visual}, where high-similarity deformations preserve palm identity while maintaining natural geometric variation.

After filtering, the remaining high-quality deformation fields form a Deformation Library $\mathcal{L} = \{\mathbf{F}_1, \mathbf{F}_2, \dots, \mathbf{F}_N\}$, which statistically represents the real-world non-rigid deformation distribution of palmprints. 
This library provides a physically meaningful geometric prior that is later sampled during generation to simulate realistic shape variations.

\subsection{Deformation-Driven Palmprint Generation}

We follow the DDIM formulation~\cite{song2021ddim} with a pretrained noise predictor $\boldsymbol{\epsilon}_\theta$ and perform sampling in three stages that progressively incorporate geometric deformation, as illustrated in~\cref{fig:total_frame}.
Let $\beta_t\!\in\!(0,1)$ be the variance schedule, $\alpha_t=1-\beta_t$, and $\bar{\alpha}_t=\prod_{s=1}^{t}\alpha_s$ with $\bar{\alpha}_0=1$.
Denote by $\mathbf{x}_t$ the intermediate variable at timestep $t$, and by $\mathbf{C}$ the current conditioning signal.
For each step $(t,t\!-\!1)$, the reverse update is
\begin{equation}
\mathbf{x}_{t-1}
=
\frac{1}{\sqrt{\alpha_t}}
\left(
\mathbf{x}_t
-
\frac{1-\alpha_t}{\sqrt{1-\bar{\alpha}_t}}\,
\boldsymbol{\epsilon}_\theta(\mathbf{x}_t, \mathbf{C}, t)
\right)
+ \sigma_t \boldsymbol{\xi},
\label{eq:ddim-eps}
\end{equation}
where $\sigma_t = \eta\sqrt{\frac{1-\bar{\alpha}_{t-1}}{1-\bar{\alpha}_t}\!\left(1-\frac{\bar{\alpha}_t}{\bar{\alpha}_{t-1}}\right)}$ and $\boldsymbol{\xi}\!\sim\!\mathcal{N}(0,\mathbf{I})$.
We use $\eta=0$ for deterministic sampling.

\paragraph{Deformed Principal Line Generation.}
Given a manual crease map $\mathbf{C}$ representing the principal-line structure of a synthetic identity, we first sample multiple deformation fields $\{\mathbf{F}_k\}$ from the same identity and warp $\mathbf{C}$ using bilinear interpolation:
\begin{equation}
\mathbf{C}^{(k)}_{w} = \mathcal{W}(\mathbf{C}, \mathbf{F}_{k}),
\end{equation}
\noindent where $\mathcal{W}(\cdot,\cdot)$ denotes bilinear sampling (cf. Eq.~\eqref{eq:warp}).

Conditioned on $\mathbf{C}_{w}^{(k)}$, the diffusion model denoises according to Eq.~\eqref{eq:ddim-eps}:
\begin{equation}
\mathbf{x}_{t-1}
=
\frac{1}{\sqrt{\alpha_t}}
\left(
\mathbf{x}_t
-
\frac{1-\alpha_t}{\sqrt{1-\bar{\alpha}_t}}\,
\boldsymbol{\epsilon}_\theta(\mathbf{x}_t, \mathbf{C}_{w}^{(k)}, t)
\right).
\label{eq:stage1}
\end{equation}

At the final step of this stage ($t^\star=0.5T$), a clean denoising step is applied to obtain the noise-free structural signal:
\begin{equation}
\mathbf{x}_{\text{clean}}
=
\frac{
\mathbf{x}_{t^\star}
-
\sqrt{1-\bar{\alpha}_{t^\star}}\,
\boldsymbol{\epsilon}_\theta(\mathbf{x}_{t^\star}, \mathbf{C}_{w}^{(k)}, t^\star)
}
{\sqrt{\bar{\alpha}_{t^\star}}}.
\label{eq:clean}
\end{equation}

The resulting clean principal-line image $\mathbf{x}_{\text{clean}}$ provides a geometry-consistent structure prior for subsequent warped noise injection.

\paragraph{Deformed Texture Generation.}

Inspired by Diff-Palm~\cite{jin2025diff}, which introduces consistency noise in the late diffusion stage, and further motivated by recent findings on noise equivariance~\cite{zhou2025alias}, we propose to inject a warped homologous Gaussian noise to add correspondingly warped textures onto the deformed principal lines. 

Unlike~\cite{jin2025diff}, which relies on DDPM and gradually aligns the noise distribution through multiple iterations, we first perform a clean denoising step in Eq.~\eqref{eq:clean} to explicitly remove the original noise component. 
This enables us to inject a new noise term in a single step and leverage deterministic DDIM sampling for acceleration, without the need to add noise at every iteration. 
An additional benefit is that the noise warping only needs to be performed once, instead of at each diffusion step, which reduces computational overhead.

Formally, the warped noise $\mathbf{n}_{\text{warp}}$ is generated from the deformation field $\mathbf{F}_k$ 
following the $\int$-Noise warping strategy~\cite{chang2024warp}. 
This noise transformation $\mathcal{T}_{\text{warp}}(\cdot)$ ensures that the warped noise maintains the Gaussian property:
\begin{equation}
\mathbf{n}_{\text{warp}} = \mathcal{T}_{\text{warp}}(\boldsymbol{\xi}, \mathbf{F}_k),
\quad \text{s.t.} \quad 
\mathbf{n}_{\text{warp}} \sim \mathcal{N}(0, \mathbf{I}).
\end{equation}

We then reconstruct the noisy state at $t^\star$ as:
\begin{equation}
\mathbf{x}_{t^\star}^{\text{new}}
=
\sqrt{\bar{\alpha}_{t^\star}}\,\mathbf{x}_{\text{clean}}
+
\sqrt{1-\bar{\alpha}_{t^\star}}\,\mathbf{n}_{\text{warp}}.
\label{eq:warp-noise}
\end{equation}

Subsequent diffusion steps continue following Eq.~\eqref{eq:stage1} to generate geometrically deformed yet identity-consistent texture aligned with the warped principal lines.

\begin{figure}
    \centering
    \includegraphics[width=0.99\linewidth]{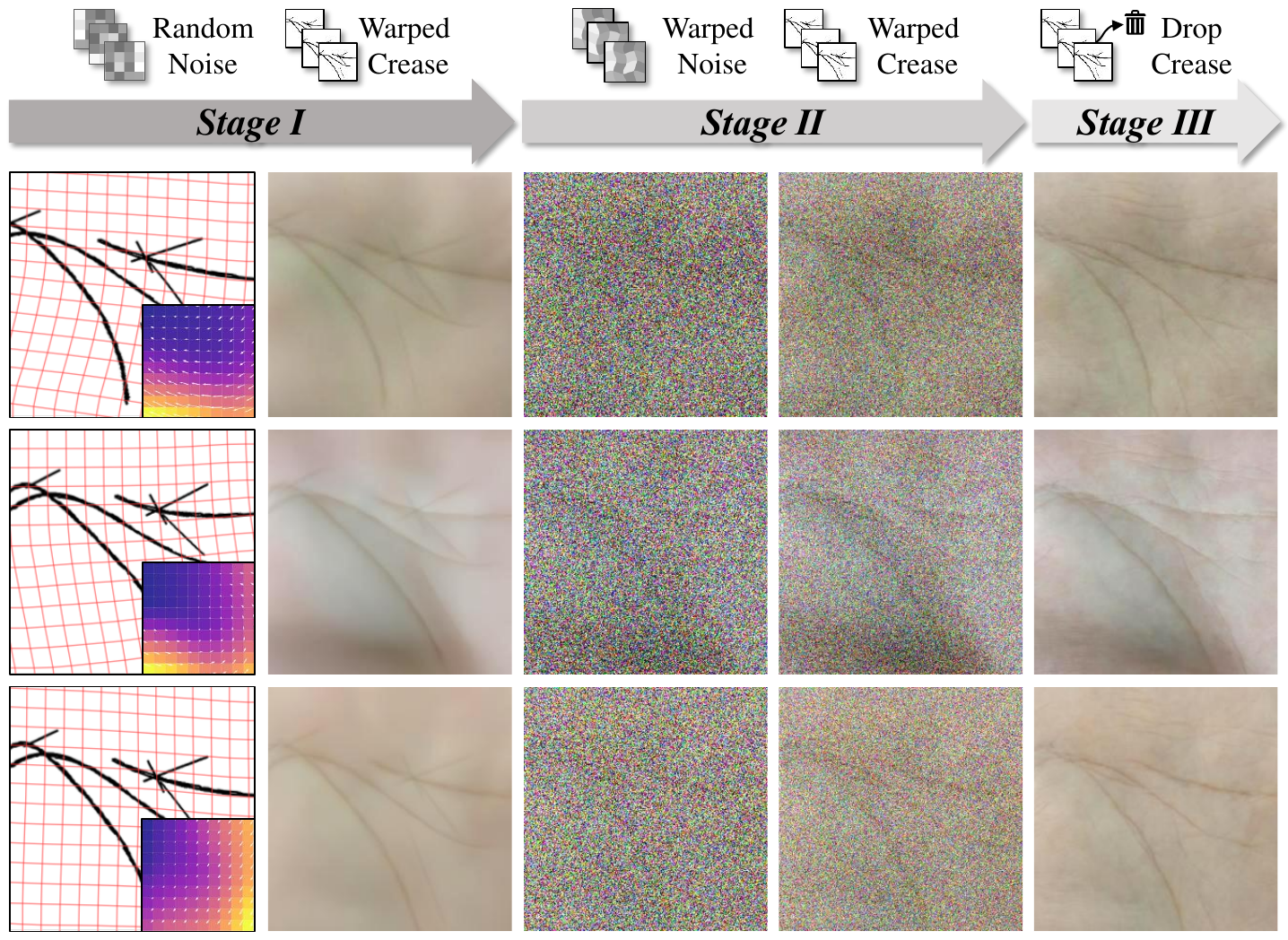}
    \caption{Visualization of the intermediate results of our deformation-driven denoising process.}
    \label{fig:sample_visual}
\end{figure}

\paragraph{Unconditional Texture Refinement.}
As illustrated in the upper-right of~\cref{fig:total_frame}, during training, a texture extractor~\cite{jin2024pce} is employed to extract palmprint textures as paired conditional inputs for the diffusion model. However, this training condition differs substantially from the manually generated crease maps used during sampling. Directly conditioning on manual crease often leads to over-constrained or unrealistic textures.

To alleviate this mismatch, we randomly drop crease conditions during training to learn an unconditional branch. During the final diffusion stage ($t < \tau_u \cdot T $), we remove the crease condition and continue denoising via the same update rule without conditioning:
\begin{equation}
\mathbf{x}_{t-1}
=
\frac{1}{\sqrt{\alpha_t}}
\left(
\mathbf{x}_t
-
\frac{1-\alpha_t}{\sqrt{1-\bar{\alpha}_t}}\,
\boldsymbol{\epsilon}_\theta(\mathbf{x}_t, \varnothing, t)
\right),
\label{eq:uncond}
\end{equation}
where $\varnothing$ denotes the absence of crease conditioning. 
After completing all steps, the final refined image is obtained as $\mathbf{x}_0$.

\cref{fig:sample_visual} visualizes the complete deformation-driven three-stage generation process, showing intermediate results of warped principal-line generation, warped texture generation, and unconditional texture refinement.

\section{Experiments}

\begin{table*}
\centering
\small
\begin{tabular}{c|c|c|c|cccccc|c}
\toprule
\multirow{2}{*}{Setting} & \multirow{2}{*}{Method} & \multirow{2}{*}{Venue} & \multirow{2}{*}{Geo Trans.} &
\multicolumn{6}{c|}{Verification Datasets} & \multirow{2}{*}{Avg} \\
\cmidrule(lr){5-10}
& & & & XJTU & MPD & TCD & CASIA & IITD & PolyU & \\
\midrule

\multirow{6}{*}{\makecell{\textbf{w/o Aug.}}} 
& RPG-Palm~\cite{shen2023rpg} & ICCV'23 & Line Osc. &  0.01 & 0.22 & 0.24 & 2.60 & 0.46 & 0.44 & 0.66\\
& PCE-Palm~\cite{jin2024pce} & AAAI'24 & Affine & 11.38 & 27.44 & 34.09 & 24.73 & \cellcolor{gray!40}\textbf{69.39} & 33.39 & 33.40 \\
& UAA~\cite{jin2025unified} & ICCV'25 & Affine & \cellcolor{gray!20}14.04 & \cellcolor{gray!20}49.55 & \cellcolor{gray!20}79.09 & \cellcolor{gray!20}55.47 & 33.75 & \cellcolor{gray!20}76.22 & \cellcolor{gray!20}51.35\\
& Diff-Palm~\cite{jin2025diff} & CVPR'25 & $\times$ & 2.52 & 3.52 & 16.08 & 4.80 & 7.58 & 24.53 & 9.84 \\
& PFIG-Palm~\cite{zou2025pfig} & TIP'25 & Line Osc. & 8.84 & 21.50 & 58.50 & 4.80 & \cellcolor{gray!20}61.45 & 66.96 & 37.01 \\
& FlowPalm (Ours) & - & Non-rigid & \cellcolor{gray!40}\textbf{17.30} & \cellcolor{gray!40}\textbf{64.29} & \cellcolor{gray!40}\textbf{91.65} & \cellcolor{gray!40}\textbf{61.07} & 30.02 & \cellcolor{gray!40}\textbf{89.35} & \cellcolor{gray!40}\textbf{58.95} \\
\cmidrule{1-11}
\multirow{6}{*}{\makecell{\textbf{w/ Aug.}}} 
& RPG-Palm~\cite{shen2023rpg} & ICCV'23 & Line Osc. & 37.51 & 38.27 & 83.73 & 50.60 & 77.33 & 92.76 & 63.37 \\
& PCE-Palm~\cite{jin2024pce} & AAAI'24 & Affine & 20.01 & 33.30 & 40.03 & 27.13 & 77.98 & 47.43 & 40.98 \\
& UAA~\cite{jin2025unified} & ICCV'25 & Affine & 55.15 & 55.80 & 91.24 & 63.33 & \cellcolor{gray!40}\textbf{88.96} & \cellcolor{gray!20}93.73 & 74.70 \\
& Diff-Palm~\cite{jin2025diff} & CVPR'25 & $\times$ & \cellcolor{gray!20}72.25 & \cellcolor{gray!20}63.09 & \cellcolor{gray!20}93.51 & \cellcolor{gray!20}66.40 & 84.55 & \cellcolor{gray!40}\textbf{94.30} & \cellcolor{gray!20}79.02 \\
& PFIG-Palm~\cite{zou2025pfig} & TIP'25 & Line Osc. & 43.06 & 40.67 & 82.20 & 47.80 & 74.45 & 83.79 & 62.00 \\
& FlowPalm (Ours) & - & Non-rigid & \cellcolor{gray!40}\textbf{86.33} & \cellcolor{gray!40}\textbf{80.37} & \cellcolor{gray!40}\textbf{97.37} & \cellcolor{gray!40}\textbf{66.53} & \cellcolor{gray!20}87.32 & 93.29 & \cellcolor{gray!40}\textbf{85.20} \\

\bottomrule
\end{tabular}
\caption{Performance analysis of TAR@FAR=$10^{-6}$ (\%) on six benchmark datasets with and without data augmentation. Geo Trans. denotes the geometric transformation type adopted in each generation method. \colorbox{gray!40}{Best} and \colorbox{gray!20}{second-best} results are highlighted.}
\label{tab:aug}
\end{table*}

\begin{table*}
\centering
\small
\begin{tabular}{c|c|c|c|cccccc|c}
\toprule
\multirow{2}{*}{Setting} & \multirow{2}{*}{Method} & \multirow{2}{*}{Venue} & \multirow{2}{*}{Geo Trans.} &
\multicolumn{6}{c|}{Verification Datasets} & \multirow{2}{*}{Avg} \\
\cmidrule(lr){5-10}
& & & & XJTU & MPD & TCD & CASIA & IITD & PolyU & \\
\midrule

\textbf{Real Only} & - & - & - & 78.05 & 73.02 & 88.70 & 32.60 & 73.54 & 95.63 & 73.59 \\
\midrule

\multirow{3}{*}{\textbf{Syn. Only}} 
 & Diff-Palm~\cite{jin2025diff} & CVPR'25 & $\times$ & \cellcolor{gray!20}72.25 & \cellcolor{gray!20}63.09 & \cellcolor{gray!20}93.51 & \cellcolor{gray!20}66.40 & \cellcolor{gray!20}84.55 & \cellcolor{gray!40}\textbf{94.30} & \cellcolor{gray!20}79.02 \\
 & PFIG-Palm~\cite{zou2025pfig} & TIP'25 & Line Osc. & 43.06 & 40.67 & 82.20 & 47.80 & 74.45 & 83.79 & 62.00 \\
 & FlowPalm (Ours) & - & Non-rigid & \cellcolor{gray!40}\textbf{86.33} & \cellcolor{gray!40}\textbf{80.37} & \cellcolor{gray!40}\textbf{97.37} & \cellcolor{gray!40}\textbf{66.53} & \cellcolor{gray!40}\textbf{87.32} & \cellcolor{gray!20}93.29 & \cellcolor{gray!40}\textbf{85.20} \\
\midrule

\multirow{3}{*}{\textbf{Syn. $+$ Real}} 
 & Diff-Palm~\cite{jin2025diff} & CVPR'25 & $\times$ & 90.05 & 84.41 & 97.42 & \cellcolor{gray!20}77.87 & \cellcolor{gray!40}\textbf{95.30} & 98.36 & 90.57 \\
 & PFIG-Palm~\cite{zou2025pfig} & TIP'25 & Line Osc. & \cellcolor{gray!20}91.01 & \cellcolor{gray!40}\textbf{89.41} & \cellcolor{gray!20}98.51 & 76.40 & \cellcolor{gray!20}95.26 & \cellcolor{gray!40}\textbf{99.03} & \cellcolor{gray!20}91.60 \\
 & FlowPalm (Ours) & - & Non-rigid & \cellcolor{gray!40}\textbf{95.28} & \cellcolor{gray!20}88.20 & \cellcolor{gray!40}\textbf{98.68} & \cellcolor{gray!40}\textbf{83.33} & 93.27 & \cellcolor{gray!20}98.92 & \cellcolor{gray!40}\textbf{92.95} \\
\midrule

\multirow{3}{*}{\textbf{Syn. $\to$ Real}} 
 & Diff-Palm~\cite{jin2025diff} & CVPR'25 & $\times$ & 89.42 & 83.70 & 97.28 & 74.40 & 96.31 & 98.90 & 90.00 \\
 & PFIG-Palm~\cite{zou2025pfig} & TIP'25 & Line Osc. & \cellcolor{gray!20}91.62 & \cellcolor{gray!20}85.30 & \cellcolor{gray!40}\textbf{99.15} & \cellcolor{gray!20}79.67 & \cellcolor{gray!20}96.70 & \cellcolor{gray!40}\textbf{99.76} & \cellcolor{gray!20}92.03 \\
 & FlowPalm (Ours) & - & Non-rigid & \cellcolor{gray!40}\textbf{91.73} & \cellcolor{gray!40}\textbf{90.44} & \cellcolor{gray!20}99.14 & \cellcolor{gray!40}\textbf{87.00} & \cellcolor{gray!40}\textbf{97.06} & \cellcolor{gray!20}99.52 & \cellcolor{gray!40}\textbf{94.15} \\

\bottomrule
\end{tabular}
\caption{Comparison of training strategies. \textbf{Syn. + Real}: mixed training; \textbf{Syn. $\to$ Real}: pre-training to fine-tuning.}
\label{tab:train_strategy}
\end{table*}

\subsection{Implementation Details}

\subsubsection{Datasets}
We conduct experiments on six benchmark palmprint databases: XJTU-UP~\cite{c42}, MPD~\cite{c44}, TCD~\cite{c45}, CASIA~\cite{c46}, IITD~\cite{c47}, and PolyU~\cite{c48}. 
For all databases, the region of interest (ROI) is extracted following the detection and cropping protocol in~\cite{c49}. Each database is evenly split by identities: the training split is used for optical flow estimation and diffusion model training, while the remaining identities for evaluating the recognition performance of generated palmprints in downstream tasks. 

\subsubsection{Deformation Prior Construction}
To construct the deformation prior, we estimate optical flow fields between real palmprint pairs using the pretrained RAFT-Large model~\cite{teed2020raft}. Each image is resized to $256\times256$ before flow estimation. We randomly sample identities that are strictly separated from the testing identities and compute up to 40 image pairs per subject. The resulting deformation fields are filtered using thresholds $\tau_d\!=\!0.01$ for discontinuity ratio and $\tau_c\!=\!0.4$ for cosine similarity.

\subsubsection{Diffusion Model Settings}
Our diffusion model adopts a U-Net backbone with five resolution levels and operates on images of size $256 \times 256$. We train with a batch size of 128 using Adam optimizer~\cite{adam} (learning rate $8\times10^{-5}$). The model is trained for 100k steps with an exponential moving average for stability. During sampling, we set the denoising process to $T=250$ timesteps, utilizing the polynomial curves proposed in~\cite{jin2025diff} as the line conditions. All diffusion model training and sampling experiments are performed on 8 NVIDIA A100 GPUs, each with 40~GB of memory.

\subsubsection{Recognition Model Settings}
To evaluate the quality of synthetic data, we train recognition models on samples generated by different methods. 
Each model is trained with 2,000 synthetic identities, and each identity contains 40 images resized to $224 \times 224$. 
The backbone is MobileFaceNet~\cite{mobileface}, optimized with SGD (momentum~0.9, weight decay~$1\times10^{-4}$) for 40 epochs under a cosine learning rate schedule (initial rate~0.02). 
Horizontal flipping is applied, effectively doubling the number of training identities. 
All recognition experiments are performed on a single NVIDIA A100 GPU.


\begin{table*}[!t]
\centering
\small
\begin{tabular}{c|
c@{\hspace{6pt}}c@{\hspace{6pt}}c@{\hspace{6pt}}c|
ccc|cc}
\toprule
\multirow{3}{*}{Method} &
\multicolumn{4}{c|}{{Configs}} &
\multicolumn{3}{c|}{{Score Distributions}} &
\multicolumn{2}{c}{{TAR@FAR=$10^{-6}\uparrow$}} \\
\cmidrule(lr){2-5} \cmidrule(lr){6-8} \cmidrule(lr){9-10}
 & \makecell[c]{Deform\\Sel.}
 & \makecell[c]{Crease\\Warp} 
 & \makecell[c]{Noise\\Warp} 
 & \makecell[c]{Drop\\Time $\tau_u$} 
 & \makecell[c]{Fréchet\\Distance$\downarrow$} 
 & \makecell[c]{Inter-cls\\Distance$\uparrow$} 
 & \makecell[c]{Intra-cls\\Distance$\downarrow$} 
 & \makecell[c]{w/o Aug.} 
 & \makecell[c]{w/ Aug.} \\
\midrule
RPG-Palm~\cite{shen2023rpg} 
& - & - & - & - 
& 0.3035 & 0.9265 & \cellcolor{gray!20}0.3341 
& 0.66 & 63.73 \\
PCE-Palm~\cite{jin2024pce} 
& - & - & - & - 
& 0.3738 & 0.8439 & 0.5657 
& 33.40 & 40.98 \\
Diff-Palm~\cite{jin2025diff} 
& - & - & - & - 
& 0.2838 & 0.9181 & \cellcolor{gray!40}\textbf{0.2592}
& 9.84 & \cellcolor{gray!20}79.02 \\
PFIG-Palm~\cite{zou2025pfig} 
& - & - & - & - 
& \cellcolor{gray!20}0.2372 & \cellcolor{gray!20}0.9425 & 0.3525 
& \cellcolor{gray!20}37.01 & 62.00 \\
FlowPalm (Ours) 
& \checkmark & \checkmark & \checkmark & 0.25 
& \cellcolor{gray!40}\textbf{0.1503} & \cellcolor{gray!40}\textbf{0.9559} & 0.3766 
& \cellcolor{gray!40}\textbf{58.95} & \cellcolor{gray!40}\textbf{85.20} \\

\midrule
\multirow{5}{*}{\shortstack{FlowPalm\\(Ours)}}
 & × & × & × & 0.25 & 0.1663 & 0.9531 & \cellcolor{gray!40}\textbf{0.2460} & 7.78 &  74.78 \\
 & × & \checkmark & \checkmark & 0.25 & 0.1545 & 0.9546 & 0.3879  & \cellcolor{gray!20}51.73 & \cellcolor{gray!20}76.61 \\
 & \checkmark & × & \checkmark & 0.25 & 0.1522 & \cellcolor{gray!20}0.9557 & 0.4256 & 27.43 & 71.00\\
& \checkmark & \checkmark & × & 0.25 & \cellcolor{gray!40}\textbf{0.1499} & 0.9556 & 0.4355 & 7.58 & 74.08  \\
& \checkmark & \checkmark & \checkmark & 0.25 & \cellcolor{gray!20}0.1503 & \cellcolor{gray!40}\textbf{0.9559} & \cellcolor{gray!20}0.3766 & \cellcolor{gray!40}\textbf{58.95} & \cellcolor{gray!40}\textbf{85.20} \\
\midrule
\multirow{5}{*}{\shortstack{FlowPalm\\(Ours)}}
 & \checkmark & \checkmark & \checkmark & 0.00 & 0.2073 & 0.9431 & \cellcolor{gray!40}\textbf{0.2972} & 29.89 & 67.25 \\
 & \checkmark & \checkmark & \checkmark & 0.10 & 0.1733 & 0.9483 & \cellcolor{gray!20}0.3343 & 39.50 & 70.58 \\
 & \checkmark & \checkmark & \checkmark & 0.20 & 0.1575 & 0.9534 & 0.3629 & 54.32 & 83.88 \\
 & \checkmark & \checkmark & \checkmark & 0.25 & \cellcolor{gray!20}0.1503 & \cellcolor{gray!20}0.9559 & 0.3766 & \cellcolor{gray!40}\textbf{58.95} & \cellcolor{gray!40}\textbf{85.20}  \\
 & \checkmark & \checkmark & \checkmark & 0.30 & \cellcolor{gray!40}\textbf{0.1433} & \cellcolor{gray!40}\textbf{0.9580} & 0.3936 & \cellcolor{gray!20}58.57 & \cellcolor{gray!20}85.14 \\
 
\bottomrule
\end{tabular}
\caption{Score distribution comparison and ablation study of the proposed FlowPalm. The top section compares FlowPalm with state-of-the-art generative methods. The middle section presents a component-wise ablation on deformation selection (Deform Sel.), crease warping (Crease Warp), and noise warping (Noise Warp). The bottom section analyzes the effect of condition drop start time $\tau_u$ in texture refinement stage.}
\label{tab:distribution_ablation}
\end{table*}

\subsection{Experimental Results}
\subsubsection{Performance Comparison and Analysis}

We compare FlowPalm with several representative palmprint generation methods, including RPG-Palm~\cite{shen2023rpg}, PCE-Palm~\cite{jin2024pce}, Diff-Palm~\cite{jin2025diff}, PFIG-Palm~\cite{zou2025pfig} and the latest augmentation method UAA~\cite{jin2025unified}. For a fair comparison, models with inconsistent training sets are retrained, while others use official weights.
The downstream recognition networks are evaluated on the six benchmark datasets, with generation performance assessed via purely synthetic training (Table~\ref{tab:aug}) and practical utility explored across diverse training paradigms (Table~\ref{tab:train_strategy}).

\textbf{Results w/ and w/o Augmentation.} As shown in Table~\ref{tab:aug}, modeling geometric variations is critical. Without augmentation, Diff-Palm~\cite{jin2025diff} struggles due to limited geometric diversity, whereas FlowPalm achieves the best performance by explicitly simulating non-rigid deformations. When applying RandAugment~\cite{cubuk2020randaugment} $(4,4)$, methods using handcrafted transformations show limited gains and may disrupt intra-class consistency. Conversely, FlowPalm effectively synergizes with augmentation, achieving the highest average TAR.

\textbf{Evaluation on Diverse Training Paradigms.} 
Table~\ref{tab:train_strategy} validates FlowPalm's practical utility. Notably, under the Syn. Only setting, models trained exclusively on our generated samples achieve 85.20\% accuracy, significantly surpassing the Real Only baseline (73.59\%) and other generative methods. This demonstrates our non-rigid deformations capture rich, realistic intra-class variations. Furthermore, seamlessly integrating FlowPalm-generated samples into mixed training (Syn. + Real) and pre-training (Syn. $\rightarrow$ Real) paradigms consistently yields state-of-the-art performance.

\subsubsection{Score Distribution Comparison and Analysis}
The upper part of Table~\ref{tab:distribution_ablation} compares score distributions using three metrics computed in a real-data embedding space: \textit{Fréchet Distance} (measuring the realism gap), \textit{Inter-class Distance} (reflecting identity discriminability), and \textit{Intra-class Distance} (evaluating intra-identity compactness). All distances are computed in the embedding space of a recognition model trained on real data.

Benefiting from the non-rigid deformation modeling and the unconditional texture refinement, our method can generate realistic geometric variations and fine-grained textures, indicating the best generation realism. 
Meanwhile, FlowPalm maintains superior identity discriminability with the highest inter-class distance, demonstrating that the non-rigid deformation does not compromise identity discriminability.

Although our intra-class distance is not the smallest, this is expected because non-rigid deformation naturally increases intra-class geometric diversity. 
However, this phenomenon is not detrimental. As further confirmed by the following ablation studies, such diversity contributes to improved robustness in downstream recognition tasks.

\begin{figure}
    \centering
    \includegraphics[width=0.95\linewidth]{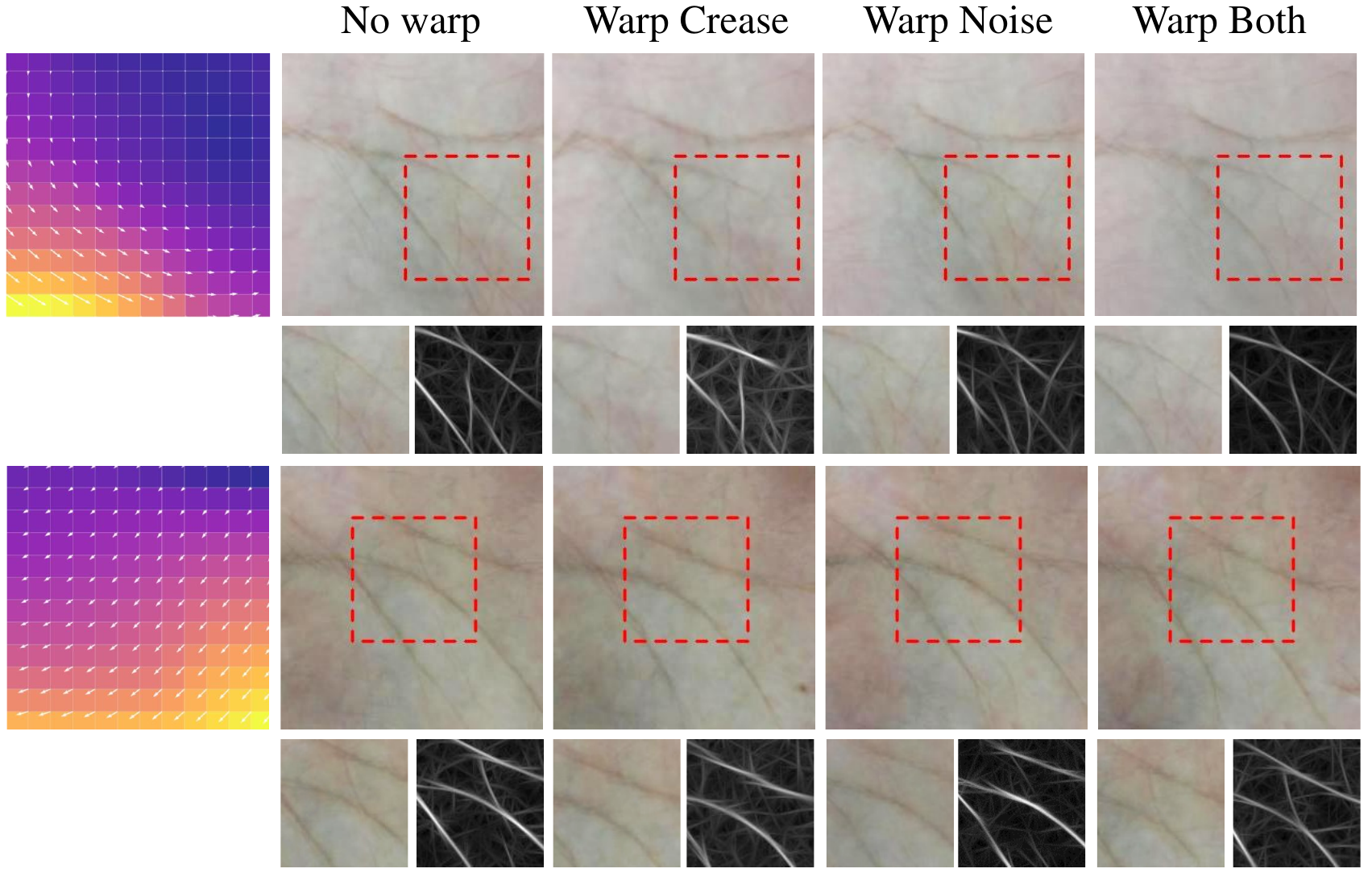}
    \caption{Visualization of the effects of crease and noise warping. Compared with the non-warped samples, warping only the crease leaves fine textures unchanged, while warping only the noise keeps the principal lines fixed. Simultaneously warping both preserves identity-consistent details that follow the deformed geometry.}
    \label{fig:ablation_warp}
\end{figure}

\subsubsection{Ablation on Deformation Components}
We analyze the contribution of each deformation-related component in the middle block of Table~\ref{tab:distribution_ablation}, where we progressively add deformation selection, crease warping, and noise warping components.

\textbf{Effect of deformation selection (Deform Sel.).} 
Applying deformation fields inherently increases geometric diversity, yielding noticeable performance gains. However, without a filtering mechanism, unselected deformations often introduce discontinuous or overly aggressive warps that severely distort the palmprint structure. Such degraded samples lead to a substantial performance degradation. This underscores the critical importance of the selection module in discarding invalid warps to preserve identity consistency while maintaining realistic diversity.

\textbf{Effect of Crease and Noise Warping.}
The middle part of Table~\ref{tab:distribution_ablation} analyzes the effects of warping the crease condition and the injected noise, with qualitative results visualized in \cref{fig:ablation_warp}. 
From the quantitative results, both crease warping and noise warping improve generation realism compared with the non-warped baseline. 
However, enabling only one of them leads to mismatch between structural deformation and texture variation, significantly increasing the intra-class distance. 
In particular, when only the noise is warped, this inconsistency heavily disrupts the recognition model, resulting in a clear performance drop. 
When both crease and noise are warped together, the alignment between structural and textural deformation is greatly enhanced, yielding the best overall performance in downstream recognition tasks.

We extract the texture response using the method~\cite{jin2024pce}. As shown in \cref{fig:ablation_warp}, compared with the non-warped samples, warping only the crease leaves fine texture details unaltered, while warping only the noise causes the principal lines to remain fixed. 
By simultaneously warping both the crease and the noise, FlowPalm produces textures whose morphology consistently follows the deformed structure.

\begin{figure}
    \centering
    \includegraphics[width=0.99\linewidth]{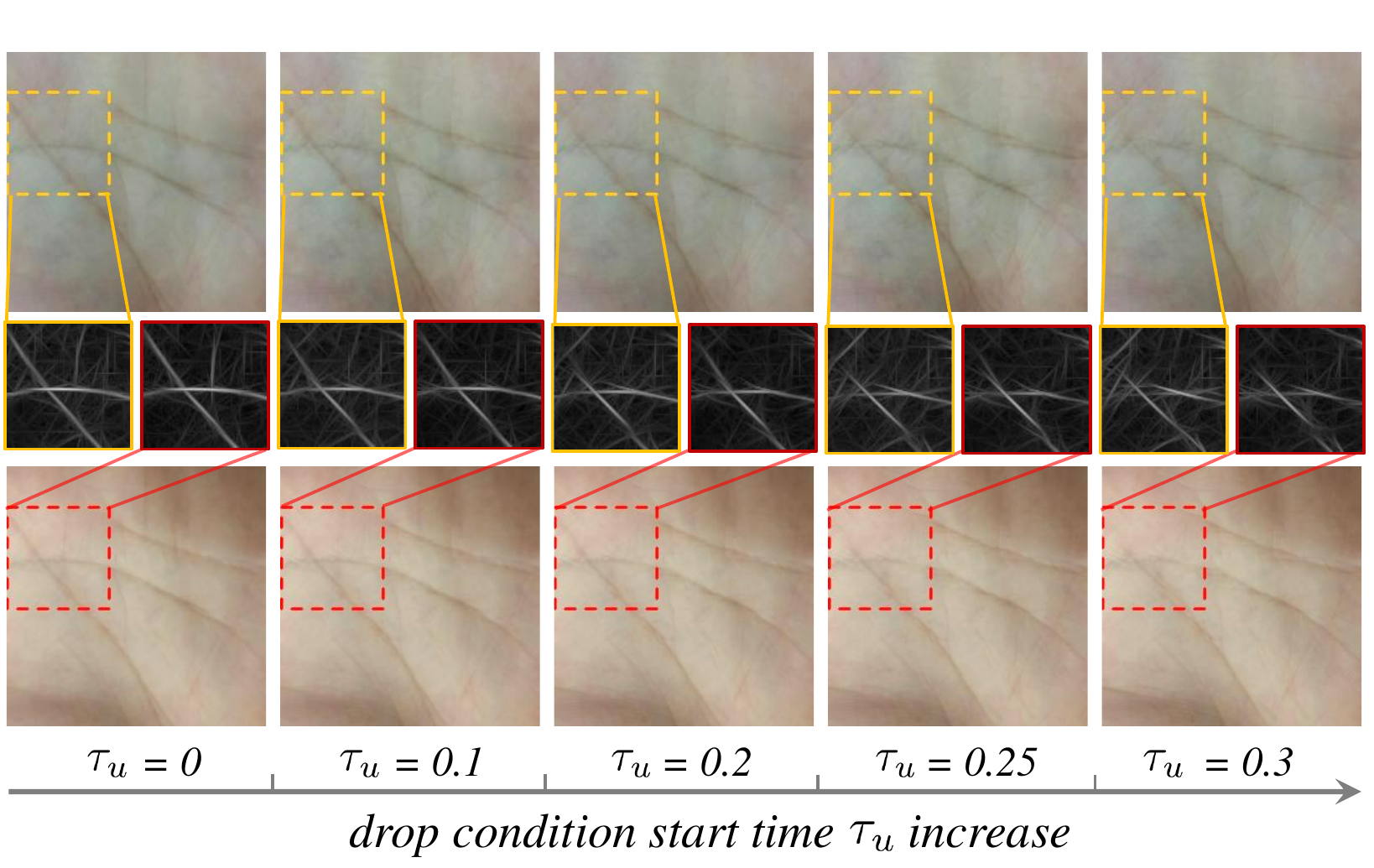}
    \caption{
    Visual effects of the condition drop start time $\tau_u$. 
    Each pair of rows shows images corresponding to the same generated identity.
    Insets highlight local texture details as $\tau_u$ increases.}
    \label{fig:ablation_t}
\end{figure}

\subsubsection{Ablation on Unconditional Refinement Time}
The lower part of Table~\ref{tab:distribution_ablation} investigates the influence of the drop start time $\tau_u$ in unconditional refinement stage. 
When $\tau_u = 0$ (no refinement), the generated distribution significantly deviates from the real data, resulting in poor performance. 
As $\tau_u$ increases, the distribution gradually aligns with real palmprints and a noticeable improvement in performance is observed. 
As shown in \cref{fig:ablation_t}, increasing $\tau_u$ progressively corrects unrealistic artificial creases (e.g., crossed principal lines) and yields richer texture details. However, excessive refinement may cause deviations in texture within identity. Setting $\tau_u = 0.25$ achieves the best balance between texture realism and identity consistency, corresponding to the optimal quantitative results.

\section{Conclusion}

In this paper, we introduced FlowPalm, an optical flow-driven framework for geometrically diverse and identity-consistent palmprint generation. By statistically analyzing optical flow fields between real palmprints, FlowPalm constructs a reliable deformation prior that captures intrinsic geometric variations. Building upon this prior, a deformation-driven denoising strategy was introduced to synthesize deformed principal lines and textures, followed by unconditional refinement for enhanced realism. 
Extensive experiments on six benchmark palmprint databases demonstrate that downstream recognition models trained exclusively on our synthesized data can surpass real data. Furthermore, integrating FlowPalm-generated samples across various training paradigms consistently yields SOTA verification performance.


\section*{Acknowledgements}
{We thank all the anonymous reviewers for their constructive comments. This work was supported in part by the National Natural Science Foundation of China (Grant Nos. 62206218 and 62376211), and in part by the Sichuan Science and Technology Program under Grant 2025ZNSFSC1494.}
{
    \small
    \bibliographystyle{ieeenat_fullname}
    \bibliography{main}

@string(CVPR= {IEEE Conf. Comput. Vis. Pattern Recog.})

@string(ICCV= {Int. Conf. Comput. Vis.})

@string(ECCV= {Eur. Conf. Comput. Vis.})

@string(ICME = {Int. Conf. Multimedia and Expo})

@string(ICIP = {IEEE Int. Conf. Image Process.})

@string(ICLR = {Int. Conf. Learn. Represent.})

@string(AAAI = {AAAI})

@string(CVPRW= {IEEE Conf. Comput. Vis. Pattern Recog. Worksh.})

@string(CVPR  = {CVPR})

@string(ICCV  = {ICCV})

@string(ECCV  = {ECCV})

@string(ICME  =	{ICME})

@string(ICIP  = {ICIP})

@string(ICLR  = {ICLR})

@string(CVPRW= {CVPRW})

@article{yang2024physics,
	title        = {Physics-driven spectrum-consistent federated learning for palmprint verification},
	author       = {Yang, Ziyuan and Teoh, Andrew Beng Jin and Zhang, Bob and Leng, Lu and Zhang, Yi},
	year         = 2024,
	journal      = {Int. J. Comput. Vis.},
	volume       = 132,
	pages        = {4253--4268}
}

@article{fan2023amgnet,
	title        = {AMGNet: Aligned multilevel gabor convolution network for palmprint recognition},
	author       = {Fan, Dandan and Liang, Xu and Zhang, Chunsheng and Jia, Wei and Zhang, David},
	year         = 2023,
	journal      = {IEEE Trans. Circuits Syst. Video Technol.},
	volume       = 34,
	pages        = {4175--4189}
}

@article{zhao2022structure,
	title        = {Structure suture learning-based robust multiview palmprint recognition},
	author       = {Zhao, Shuping and Fei, Lunke and Wen, Jie and Zhang, Bob and Zhao, Pengyang and Li, Shuyi},
	year         = 2022,
	journal      = {IEEE Trans. Neural Networks Learn. Syst.},
	volume       = 35,
	pages        = {8401--8413}
}

@article{liu2025sf2net,
	title        = {SF2Net: Sequence Feature Fusion Network for Palmprint Verification},
	author       = {Yunlong Liu and Lu Leng and Ziyuan Yang and Andrew Beng Jin Teoh and Bob Zhang},
	year         = 2025,
	journal      = {IEEE Trans. Inf. Forensics Secur.},
	volume       = 20,
	pages        = {9936--9949}
}

@article{pan2025hierarchical,
	title        = {Hierarchical Cross-Modal Image Generation for Multimodal Biometric Recognition With Missing Modality},
	author       = {Zaiyu Pan and Shuangtian Jiang and Xiao Yang and Hai Yuan and Jun Wang},
	year         = 2025,
	journal      = {{IEEE} Trans. Inf. Forensics Secur.},
	volume       = 20,
	pages        = {4308--4321}
}

@article{shao2024generating,
	title        = {Generating Stylized Features for Single-Source Cross-Dataset Palmprint Recognition With Unseen Target Dataset},
	author       = {Huikai Shao and Pengxu Li and Dexing Zhong},
	year         = 2024,
	journal      = {{IEEE} Trans. Image Process.},
	volume       = 33,
	pages        = {4911--4922}
}

@article{liu2022data,
	title        = {Data Protection in Palmprint Recognition via Dynamic Random Invisible Watermark Embedding},
	author       = {Chengcheng Liu and Dexing Zhong and Huikai Shao},
	year         = 2022,
	journal      = {{IEEE} Trans. Circuits Syst. Video Technol.},
	volume       = 32,
	pages        = {6927--6940}
}

@article{goodfellow2020gan,
	title        = {Generative adversarial networks},
	author       = {Goodfellow, Ian and Pouget-Abadie, Jean and Mirza, Mehdi and Xu, Bing and Warde-Farley, David and Ozair, Sherjil and Courville, Aaron and Bengio, Yoshua},
	year         = 2020,
	journal      = {Commun. ACM},
	volume       = 63,
	pages        = {139--144}
}

@article{ho2020denoising,
	title        = {Denoising diffusion probabilistic models},
	author       = {Ho, Jonathan and Jain, Ajay and Abbeel, Pieter},
	year         = 2020,
	journal      = {Adv. Neural Inf. Process. Syst. (NeurIPS)},
	volume       = 33,
	pages        = {6840--6851}
}

@inproceedings{zhao2022bezierpalm,
	title        = {B{\'e}zierpalm: A free lunch for palmprint recognition},
	author       = {Zhao, Kai and Shen, Lei and Zhang, Yingyi and Zhou, Chuhan and Wang, Tao and Zhang, Ruixin and Ding, Shouhong and Jia, Wei and Shen, Wei},
	year         = 2022,
	booktitle    = {Proc. Eur. Conf. Comput. Vis. (ECCV)},
	pages        = {19--36}
}

@article{Zhong2025RegPalm,
	title        = {RegPalm: Toward Large-Scale Open-Set Palmprint Recognition by Reducing Pattern Variance},
	author       = {Yaoyao Zhong and Weilong Chai and Libin Wang and Dandan Zheng and Huiyuan Fu and Huadong Ma},
	year         = 2025,
	journal      = {IEEE Trans. Inf. Forensics Secur.},
	volume       = 20,
	pages        = {8541--8554}
}

@inproceedings{jin2025diff,
	title        = {Diff-Palm: Realistic Palmprint Generation with Polynomial Creases and Intra-Class Variation Controllable Diffusion Models},
	author       = {Jin, Jianlong and Zhao, Chenglong and Zhang, Ruixin and Shang, Sheng and Xu, Jianqing and Zhang, Jingyun and Wang, ShaoMing and Zhao, Yang and Ding, Shouhong and Jia, Wei and others},
	year         = 2025,
	booktitle    = {Proc. IEEE/CVF Conf. Comput. Vis. Pattern Recognit. (CVPR)},
	pages        = {26367--26376}
}

@inproceedings{shen2023rpg,
	title        = {Rpg-palm: Realistic pseudo-data generation for palmprint recognition},
	author       = {Shen, Lei and Jin, Jianlong and Zhang, Ruixin and Li, Huaen and Zhao, Kai and Zhang, Yingyi and Zhang, Jingyun and Ding, Shouhong and Zhao, Yang and Jia, Wei},
	year         = 2023,
	booktitle    = {Proc. IEEE/CVF Int. Conf. Comput. Vis. (ICCV)},
	pages        = {19605--19616}
}

@article{zou2025pfig,
	title        = {PFIG-Palm: Controllable Palmprint Generation via Pixel and Feature Identity Guidance},
	author       = {Zou, Yuchen and Shao, Huikai and Liu, Chengcheng and Zhu, Siyu and Hou, Zongqing and Zhong, Dexing},
	year         = 2025,
	journal      = {IEEE Trans. Image Process.},
	volume       = 34,
	pages        = {6603--6615}
}

@inproceedings{jin2024pce,
	title        = {Pce-palm: Palm crease energy based two-stage realistic pseudo-palmprint generation},
	author       = {Jin, Jianlong and Shen, Lei and Zhang, Ruixin and Zhao, Chenglong and Jin, Ge and Zhang, Jingyun and Ding, Shouhong and Zhao, Yang and Jia, Wei},
	year         = 2024,
	booktitle    = {Proc. AAAI Conf. Artif. Intell. (AAAI)},
	pages        = {2616--2624}
}

@inproceedings{teed2020raft,
	title        = {Raft: Recurrent all-pairs field transforms for optical flow},
	author       = {Teed, Zachary and Deng, Jia},
	year         = 2020,
	booktitle    = {Proc. Eur. Conf. Comput. Vis. (ECCV)},
	pages        = {402--419}
}

@article{joshi2024synthetic,
	title        = {Synthetic data in human analysis: A survey},
	author       = {Joshi, Indu and Grimmer, Marcel and Rathgeb, Christian and Busch, Christoph and Bremond, Francois and Dantcheva, Antitza},
	year         = 2024,
	journal      = {IEEE Trans. Pattern Anal. Mach. Intell.},
	volume       = 46,
	pages        = {4957--4976}
}

@inproceedings{papantoniou2024arc2face,
	title        = {Arc2face: A foundation model for id-consistent human faces},
	author       = {Papantoniou, Foivos Paraperas and Lattas, Alexandros and Moschoglou, Stylianos and Deng, Jiankang and Kainz, Bernhard and Zafeiriou, Stefanos},
	year         = 2024,
	booktitle    = {Proc. Eur. Conf. Comput. Vis. (ECCV)},
	pages        = {241--261}
}

@inproceedings{xu2024id,
	title        = {ID\({}^{\mbox{3}}\): Identity-Preserving-yet-Diversified Diffusion Models for Synthetic Face Recognition},
	author       = {Jianqing Xu and Shen Li and Jiaying Wu and Miao Xiong and Ailin Deng and Jiazhen Ji and Yuge Huang and Guodong Mu and Wenjie Feng and Shouhong Ding and Bryan Hooi},
	year         = 2024,
	booktitle    = {Adv. Neural Inf. Process. Syst. (NeurIPS)}
}

@inproceedings{boutros2023idiff,
	title        = {Idiff-face: Synthetic-based face recognition through fizzy identity-conditioned diffusion model},
	author       = {Boutros, Fadi and Grebe, Jonas Henry and Kuijper, Arjan and Damer, Naser},
	year         = 2023,
	booktitle    = {Proc. IEEE/CVF Int. Conf. Comput. Vis. (ICCV)},
	pages        = {19650--19661}
}

@inproceedings{wang2025facea,
	title        = {FaceA-Net: Facial Attribute-Driven ID Preserving Image Generation Network},
	author       = {Wang, Jiayu and Yu, Yue and Chen, Jingjing and Dai, Qi and Jiang, Yu-Gang},
	year         = 2025,
	booktitle    = {Proc. AAAI Conf. Artif. Intell. (AAAI)},
	pages        = {7736--7743}
}

@inproceedings{deng2019arcface,
	title        = {Arcface: Additive angular margin loss for deep face recognition},
	author       = {Deng, Jiankang and Guo, Jia and Xue, Niannan and Zafeiriou, Stefanos},
	year         = 2019,
	booktitle    = {Proc. IEEE/CVF Conf. Comput. Vis. Pattern Recognit. (CVPR)},
	pages        = {4690--4699}
}

@inproceedings{kim2022adaface,
	title        = {Adaface: Quality adaptive margin for face recognition},
	author       = {Kim, Minchul and Jain, Anil K and Liu, Xiaoming},
	year         = 2022,
	booktitle    = {Proc. IEEE/CVF Conf. Comput. Vis. Pattern Recognit. (CVPR)},
	pages        = {18750--18759}
}

@inproceedings{shao2019palmgan,
	title        = {PalmGAN for cross-domain palmprint recognition},
	author       = {Shao, Huikai and Zhong, Dexing and Li, Yuhan},
	year         = 2019,
	booktitle    = {Proc. IEEE Int. Conf. Multimedia Expo (ICME)},
	pages        = {1390--1395}
}

@article{shao2021jpfa,
	title        = {Towards cross-dataset palmprint recognition via joint pixel and feature alignment},
    author={Shao, Huikai and Zhong, Dexing},
	year         = 2021,
	journal      = {IEEE Trans. Image Process.},
	volume       = 30,
	pages        = {3764--3777}
}

@article{zhu2023contactless,
	title        = {Contactless palmprint image recognition across smartphones with self-paced CycleGAN},
	author       = {Zhu, Qi and Xin, Guangnan and Fei, Lunke and Liang, Dong and Zhang, Zheng and Zhang, Daoqiang and Zhang, David},
	year         = 2023,
	journal      = {IEEE Trans. Inf. Forensics Secur.},
	volume       = 18,
	pages        = {4944--4954}
}

@article{dhariwal2021diffusion,
	title        = {Diffusion models beat gans on image synthesis},
	author       = {Dhariwal, Prafulla and Nichol, Alexander},
	year         = 2021,
	journal      = {Adv. Neural Inf. Process. Syst. (NeurIPS)},
	volume       = 34,
	pages        = {8780--8794}
}

@inproceedings{nichol2022glide,
	title        = {{GLIDE:} Towards Photorealistic Image Generation and Editing with Text-Guided Diffusion Models},
	author       = {Alexander Quinn Nichol and Prafulla Dhariwal and Aditya Ramesh and Pranav Shyam and Pamela Mishkin and Bob McGrew and Ilya Sutskever and Mark Chen},
	year         = 2022,
	booktitle    = {Proc. Int. Conf. Mach. Learn. (ICML)},
	pages        = {16784--16804}
}

@inproceedings{rombach2022ldm,
	title        = {High-resolution image synthesis with latent diffusion models},
	author       = {Rombach, Robin and Blattmann, Andreas and Lorenz, Dominik and Esser, Patrick and Ommer, Bj{\"o}rn},
	year         = 2022,
	booktitle    = {Proc. IEEE/CVF Conf. Comput. Vis. Pattern Recognit. (CVPR)},
	pages        = {10684--10695}
}

@inproceedings{zhang2023controlnet,
	title        = {Adding conditional control to text-to-image diffusion models},
	author       = {Zhang, Lvmin and Rao, Anyi and Agrawala, Maneesh},
	year         = 2023,
	booktitle    = {Proc. IEEE/CVF Int. Conf. Comput. Vis. (ICCV)},
	pages        = {3836--3847}
}

@inproceedings{koley2024sketch,
	title        = {It's All About Your Sketch: Democratising Sketch Control in Diffusion Models},
	author       = {Koley, Subhadeep and Bhunia, Ayan Kumar and Sekhri, Deeptanshu and Sain, Aneeshan and Chowdhury, Pinaki Nath and Xiang, Tao and Song, Yi-Zhe},
	year         = 2024,
	booktitle    = {Proc. IEEE/CVF Conf. Comput. Vis. Pattern Recognit. (CVPR)},
	pages        = {7204--7214}
}

@inproceedings{koley2024text,
	title        = {Text-to-image diffusion models are great sketch-photo matchmakers},
	author       = {Koley, Subhadeep and Bhunia, Ayan Kumar and Sain, Aneeshan and Chowdhury, Pinaki Nath and Xiang, Tao and Song, Yi-Zhe},
	year         = 2024,
	booktitle    = {Proc. IEEE/CVF Conf. Comput. Vis. Pattern Recognit. (CVPR)},
	pages        = {16826--16837}
}

@inproceedings{chang2024warp,
	title        = {How {I} Warped Your Noise: a Temporally-Correlated Noise Prior for Diffusion Models},
	author       = {Pascal Chang and Jingwei Tang and Markus Gross and Vinicius C. Azevedo},
	year         = 2024,
	booktitle    = {Proc. Int. Conf. Learn. Represent. (ICLR)}
}

@inproceedings{zhou2025alias,
	title        = {Alias-free latent diffusion models: Improving fractional shift equivariance of diffusion latent space},
	author       = {Zhou, Yifan and Xiao, Zeqi and Yang, Shuai and Pan, Xingang},
	year         = 2025,
	booktitle    = {Proc. IEEE/CVF Conf. Comput. Vis. Pattern Recognit. (CVPR)},
	pages        = {34--44}
}

@inproceedings{burgert2025go,
	title        = {Go-with-the-flow: Motion-controllable video diffusion models using real-time warped noise},
	author       = {Burgert, Ryan and Xu, Yuancheng and Xian, Wenqi and Pilarski, Oliver and Clausen, Pascal and He, Mingming and Ma, Li and Deng, Yitong and Li, Lingxiao and Mousavi, Mohsen and others},
	year         = 2025,
	booktitle    = {Proc. IEEE/CVF Conf. Comput. Vis. Pattern Recognit. (CVPR)},
	pages        = {13--23}
}

@inproceedings{song2021ddim,
	title        = {Denoising Diffusion Implicit Models},
	author       = {Jiaming Song and Chenlin Meng and Stefano Ermon},
	year         = 2021,
	booktitle    = {Proc. Int. Conf. Learn. Represent. (ICLR)}
}

@article{c42,
	title        = {Deep Distillation Hashing for Unconstrained Palmprint Recognition},
	author       = {Huikai Shao and Dexing Zhong and Xuefeng Du},
	year         = 2021,
	journal      = {IEEE Trans. Instrum. Meas.},
	volume       = 70,
	pages        = {1--13}
}

@article{c44,
	title        = {Towards palmprint verification on smartphones},
	author       = {Zhang, Yingyi and Zhang, Lin and Zhang, Ruixin and Li, Shaoxin and Li, Jilin and Huang, Feiyue},
	year         = 2020,
	journal      = {arXiv preprint arXiv:2003.13266}
}

@article{c45,
	title        = {Towards contactless palmprint recognition: A novel device, a new benchmark, and a collaborative representation based identification approach},
	author       = {Zhang, Lin and Li, Lida and Yang, Anqi and Shen, Ying and Yang, Meng},
	year         = 2017,
	journal      = {Pattern Recognit.},
	volume       = 69,
	pages        = {199--212}
}

@inproceedings{c46,
	title        = {Multispectral palm image fusion for accurate contact-free palmprint recognition},
	author       = {Hao, Ying and Sun, Zhenan and Tan, Tieniu and Ren, Chao},
	year         = 2008,
	booktitle    = {Proc. IEEE Int. Conf. Image Process. (ICIP)},
	pages        = {281--284}
}

@article{c47,
	title        = {Personal identification using multibiometrics rank-level fusion},
	author       = {Kumar, Ajay and Shekhar, Sumit},
	year         = 2010,
	journal      = {IEEE Trans. Syst. Man Cybern.},
	volume       = 41,
	pages        = {743--752}
}

@article{c48,
	title        = {An online system of multispectral palmprint verification},
	author       = {Zhang, David and Guo, Zhenhua and Lu, Guangming and Zhang, Lei and Zuo, Wangmeng},
	year         = 2009,
	journal      = {IEEE Trans. Instrum. Meas.},
	volume       = 59,
	pages        = {480--490}
}

@inproceedings{c49,
	title        = {Pay by showing your palm: A study of palmprint verification on mobile platforms},
	author       = {Zhang, Yingyi and Zhang, Lin and Liu, Xiao and Zhao, Shengjie and Shen, Ying and Yang, Yukai},
	year         = 2019,
	booktitle    = {Proc. IEEE Int. Conf. Multimedia Expo (ICME)},
	pages        = {862--867}
}

@inproceedings{mobileface,
	title        = {Mobilefacenets: Efficient cnns for accurate real-time face verification on mobile devices},
	author       = {Chen, Sheng and Liu, Yang and Gao, Xiang and Han, Zhen},
	year         = 2018,
	booktitle    = {Proc. Chin. Conf. Biometric Recognit. (CCBR)},
	pages        = {428--438}
}

@inproceedings{adam,
	title        = {Adam: {A} Method for Stochastic Optimization},
	author       = {Diederik P. Kingma and Jimmy Ba},
	year         = 2015,
	booktitle    = {Proc. Int. Conf. Learn. Represent. (ICLR)}
}

@inproceedings{yan2025consistent,
  title={Consistent Flow Distillation for Text-to-3D Generation},
  author={Yan, Runjie and Chen, Yinbo and Wang, Xiaolong},
  year         = 2025,
  booktitle    = {Proc. Int. Conf. Learn. Represent. (ICLR)}
}

@inproceedings{muller2024multidiff,
  title={Multidiff: Consistent novel view synthesis from a single image},
  author={M{\"u}ller, Norman and Schwarz, Katja and R{\"o}ssle, Barbara and Porzi, Lorenzo and Bulo, Samuel Rota and Nie{\ss}ner, Matthias and Kontschieder, Peter},
  booktitle    = {Proc. IEEE/CVF Conf. Comput. Vis. Pattern Recognit. (CVPR)},
  pages={10258--10268},
  year={2024}
}

@article{liu2025equivdm,
  title={EquiVDM: Equivariant Video Diffusion Models with Temporally Consistent Noise},
  author={Liu, Chao and Vahdat, Arash},
  journal={arXiv preprint arXiv:2504.09789},
  year={2025}
}

@inproceedings{cubuk2020randaugment,
  title={Randaugment: Practical automated data augmentation with a reduced search space},
  author={Cubuk, Ekin D and Zoph, Barret and Shlens, Jonathon and Le, Quoc V},
  booktitle = {Proc. IEEE/CVF Conf. Comput. Vis. Pattern Recognit. Workshops (CVPRW)},
  pages={702--703},
  year={2020}
}

@inproceedings{jin2025unified,
  title={Unified Adversarial Augmentation for Improving Palmprint Recognition},
  author={Jin, Jianlong and Zhao, Chenglong and Zhang, Ruixin and Shang, Sheng and Zhao, Yang and Wang, Jun and Zhang, Jingyun and Ding, Shouhong and Jia, Wei and Wu, Yunsheng},
  booktitle = {Proc. IEEE/CVF Int. Conf. Comput. Vis. (ICCV)},
  pages={14141--14151},
  year={2025}
}
}


\end{document}